\documentclass[letterpaper, 10 pt, conference]{ieeeconf}  % Comment this line out if you need a4paper

\IEEEoverridecommandlockouts                              % This command is only needed if 
\usepackage{listings}
\usepackage{graphicx}
\usepackage{multirow}
\usepackage{amsmath}
\usepackage{url}
\usepackage{pgfplots}
\pgfplotsset{width=10cm,compat=1.9}

\title{\LARGE \bf Advanced Lane Detection Model for the Virtual Development of Highly Automated Functions*}
\author{Philip Pannagger$^{1}$, Demin Nalic$^{2}$, Faris Orucevic$^{3}$, Arno Eichberger$^{4}$, Branko Rogic$^{5}$% Ger  <-this % stops a space
\thanks{*This work is founded by the Austrian Federal Ministry of Transport, Innovation and Technology as part of the FFG Program "EFREtop"}
\thanks{P. Pannagger$^{1}$, D.Nalic,$^{2}$, A.Eichberger$^{3}$ F. Orucevic$^{4}$ are with the Automotive Engineering Department, Graz University of Technology, Austria. pannagger@student.tugraz.at, demin.nalic@tugraz.at, faris.orucevic@tugraz.at, arno.eichberger@tugraz.at}
\thanks{Branko Rogic$^{5}$ is with ADAS Simulation Department at MAGNA Steyr Fahrzeugtechnik AG \& CO KG. branko.rogic@magna.com}}
\begin{document}

\begin{figure*}[!t]\
	\begin{center} {This work has been submitted to the IEEE for possible publication. Copyright may be transferred without notice, after which this version may no longer be accessible.
		} 
	\end{center}
\end{figure*}
\newpage

\maketitle
\thispagestyle{empty}
\pagestyle{empty}

%%%%%%%%%%%%%%%%%%%%%%%%%%%%%%%%%%%%%%%%%%%%%%%%%%%%%%%%%%%%%%%%%%%%%%%%%%%%%%%%
\begin{abstract}

Virtual development and prototyping has already become an integral part in the field of automated driving systems (ADS). There are plenty of software tools that are used for the virtual development of ADS. One such tool is CarMaker from IPG Automotive, which is widely used in the scientific community and in the automotive industry. It offers a broad spectrum of implementation and modelling possibilities of the vehicle, driver behavior, control, sensors, and environmental models. Focusing on the virtual development of highly automated driving functions on the vehicle guidance level, it is essential to perceive the environment in a realistic manner. For the longitudinal and lateral path guidance line detection sensors are necessary for the determination of the relevant perceiving vehicle and for the planning of trajectories. For this purpose, a lane sensor model was developed in order to efficiently detect lanes in the simulation environment of CarMaker. The so-called advanced lane detection model (ALDM) is optimized regarding the calculation time and is for the lateral and longitudinal vehicle guidance in CarMaker. 

\end{abstract}

\begin{keywords}
Simulation and Modeling; Driver Assistance Systems; Sensing, Vision, and Perception
\end{keywords}

%%%%%%%%%%%%%%%%%%%%%%%%%%%%%%%%%%%%%%%%%%%%%%%%%%%%%%%%%%%%%%%%%%%%%%%%%%%%%%%%

\section{INTRODUCTION}

Speaking of highly ADS on the vehicle guidance level, developers and the researchers face different tasks regarding the control, perception and path planning units. One such system, is for example the highway chauffeur (HWC). One of the main capabilities of a HWC is automated vehicle guidance in lateral and longitudinal direction on highways. Functional requirements and constraints of a HWC system are comprehensively defined within the PEGASUS project \cite{pegasus}. The HWC combines four automated driving functions (ADF), the ACC and AEB function for longitudinal control and LCA and LKA function for the lateral control. An essential part of these ADFs is the perception of the environment which contains a variety of static or dynamic objects like pedestrians, roads, road objects, traffic participants etc.. Virtual development and testing complex ADS as the HWC is, requires complex virtual environments with realistic models \cite{req_1}-\cite{req_3}. Essential environment model for all ADFs is the lane detection sensor model which is needed for the lane detection unit of a certain ADF. In particular, the lane detection is a significant subject regarding the lane keeping assistant (LKA) and the lane change assistant (LCA), see \cite{use_1}-\cite{use_5}. Many calculations and decisions like the driving lane computation or target detection, are based on the result of the detected lane markings (LM) or road boundaries. The LM on the left and on the right side of are crucial for the calculation of the optimized driving line positioned on the actual driving lane used by the LKA. For simply keeping the lane the preview distance of the lane detection has not to be very high, H\"ober \cite{c2} describes a LKA algorithm which uses a preview distance with about 10 m. For the LKA it is sufficient to know the course of the road for the few meters in order to follow it, however for a HWC the target detection needs to know the course of the road for a higher distance to distinguish if a target is on the currently driving lane or not. The lane detection therefore has to work properly for larger distances. This paper describes a lane detection which can be used for the simulation environments of CarMaker but also as a concept for other virtual Environments. In this work the focus is on the CarMaker line detection model and the model which was presented in \cite{c2}. After introducing the cloud of points measured by the line sensor model provided by IPG CarMaker in section~\ref{sec:lineDetection} and several issues related to this model in section~\ref{section:Problem} a the ALDM is presented. The ALDM offers a fast and reliable solution used in ADFs. Based on the LKA and LCA, simulation results will be shown in order to demonstrate the functionality of this sensor. The CarMaker 8.1.1 Version was used and the Simulink interface to CarMaker of MATLAB 2018b.

% Lane Detection Models 

% Advanced Lane Detection Model
	% Use case Lane Keeping
	% Lane Change Planing
	
% Simulation Results

% Conclusion

%%%%%%%%%%%%%%%%%%%%%%%%%%%%%%%%%%%%%%%%%%%%%%%%%%%%%%%%%%%%%%%%%%%%%%%%%%%%%%%%
\section{Line Sensor Model}
\label{sec:lineDetection}
\begin{figure}[t]
  \centering
     \includegraphics[width=0.48\textwidth]{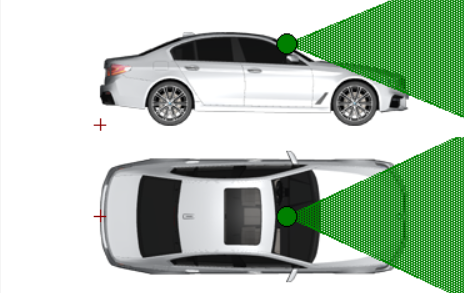}
  \caption{Line Sensor position, orientation and angle of view}
  \label{fig:LKACamera}
\end{figure}

Depending on the ADF function for the vehicle guidance the line detection unit can have different purposes. For the longitudinal control of a vehicle under test (VUT) equipped whit e.g. an ACC, the line detection is used to determine weather a vehicle in front is on the VUT line or not. For the lateral control systems e.g. a LKA or LCA the line detection is used to determine the desired line for the trajectory calculation and planing. Using the virtual environment in CarMaker for ADS development a camera or line sensor model which is provided. This camera is positioned on a VUT as depicted in figure ~\ref{fig:LKACamera}. The range of view in driving direction is about 200 m, as described and used in~\cite{c8}.
The line sensor provides following output signal, as described in the user guide~\cite{c1}:
\begin{itemize}
    \item Number of lines detected at left and right
    \item User defined color index of detected line left/right (only for sensor, independent of color displayed in IPGMovie)
    \item Height (used for traffic barriers)
    \item Line type:
    \begin{itemize}
        \item 1 = continuous single line
        \item 2 = dashed line
        \item 3 = dotted line
        \item ...
    \end{itemize}
    \item Additional external rotation of sensor, expressed in mounted frame
    \item Additional external travel of sensor, expressed in mounted frame
    \item Time stamp of updated sensor signals
    \item Point list with 30 items for both the relevant lines on right and left side
    \item Extended point list for all detected lines (c-code only)
\end{itemize}
For roads created with the Scenario/Road editor provided by CarMaker the picture processing unit (PCU) of the camera model detects every LM as a single object, no matter if it is a continuous single line, a dashed line or an other type. 
\begin{figure}[ht]
  \centering
     \includegraphics[width=0.48\textwidth]{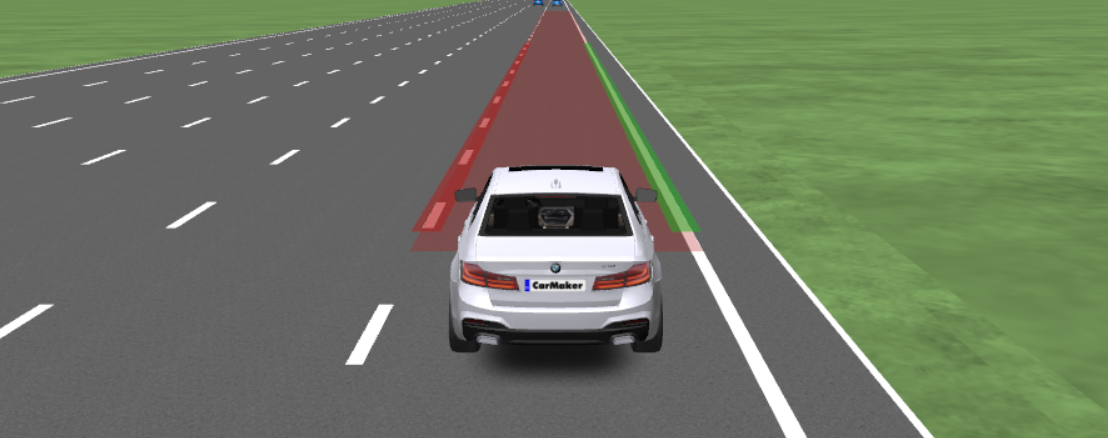}
  \caption{line detection example using a simple road created with the editor tool provided by IPG}
  \label{fig:LKALineDetecion}
\end{figure}
Figure ~\ref{fig:LKALineDetecion} shows a scenario where a dashed line presents the left LM of the current driving line and a continuous single line on the right side. Each of them are detected as one line-object, shown by the red and green overlay.\\
The CarMaker line sensor module categorizes detected lines into two groups, the lines on the right and the lines on the left side of the ego vehicle. All values are based on their relative distance to the middle axis of the car. Is the beginning of the line on right side of the middle axis it is assigned to the right side, otherwise to the left. For the implementation of a ADF the data provided by CarMaker without Simulink is not sufficient, because it does not offer access to the coordinates of the detected cloud of points representing the LMs. This extended point list is important to calculated the driving trajectory. Therefore, a CarMaker C-code interface allows the possibility to read and extend a point list of the LM and process the data which is forwarded to the Simulink simulation.\\
The extended point list for the data measured by the line sensor is stored in the variable \textit{LineSensor} of type \textit{tLineSensor}. It can be accessed via C code like shown in table~\ref{listing:listingCVariables} when including the respective files.
\begin{table}[h]
    \begin{lstlisting}
    #include <CarMaker.h>
    #include "Vehicle/Sensor_Line.h"    
        LineSensor[m].LLines.L[n].ds[p][q]
        LineSensor[m].RLines.L[n].ds[p][q]
    \end{lstlisting}
    \begin{tabular}{lll}
        $m_{l}$&...&index of the line sensor\\
        n&...&index of the detected line\\
        p&...&index of the point on the detected line\\
        q&...&0 for x-value, 1 for y-value\\
    \end{tabular}
    \caption{Listing c variables for reading extended point list}
    \label{listing:listingCVariables}
\end{table}
Using the C-code interface of CarMaker it was possible to install multiple line sensors on one car. The index \textit{m} specifies the line sensor number which is installed and accessed via the interface. For the implementation of an longitudinal and lateral control algorithm two line sensors can be used, one forward and one backward looking sensor. If \textit{m} is 0 the forward looking line sensor is accessed, if it is 1 the backward looking line sensor is chosen.\\
Every line sensor is able to detect multiple line-objects at a time. To choose one of these objects, the index \textit{n} defines which one is selected. How many lines are actually detected is stored in: 
\begin{lstlisting}
LineSensor[m].LLines.nLine
\end{lstlisting}
The array size used for \textit{L[n]} of 100 presents the upper limit. The variable \textit{ds} can be interpreted as delta stretch. It stores the relative distance (x,y and z direction) between the sensor and each point on the line. The value \textit{p} indicates a specific point in the vector. CarMaker sets an upper limit of 200 for the number of points used per line. In practice \textit{p} can be expected in a maximum range from 0 to $\lfloor\frac{\textit{ld\_range}}{2}\rfloor$, where \textit{ld\_range} is the vision range of the camera. The value \textit{p} = 0 accesses the nearest simulated point with the relative distance 0 m (0 means in this case the start of the sensors field of view), \textit{p} = 1 accesses the next nearest simulated point with the relative distance of about 2 m, \textit{p} = 2 represent 4 m and so on. Every increment of \textit{p} means an increment of approximately 2 m in x-direction.  With the index \textit{q} the respective coordinate is chosen, $q=0$ is the exact x coordinate, $q=1$ the y coordinate and $q=3$ the z coordinate. The z coordinate of the point is nearly zero for every point, as it lies on the street, and apart from this it is neither noticeable nor interesting. Under-passing another street is the only case when the z-coordinate may be important important, as in the simulation the lines on the street above may be noticed by the sensor, but that can only happen in simulation.
\section{Lane Detection Issues} \label{section:Problem}
\begin{figure}[h]
  \centering
     \includegraphics[width=0.48\textwidth]{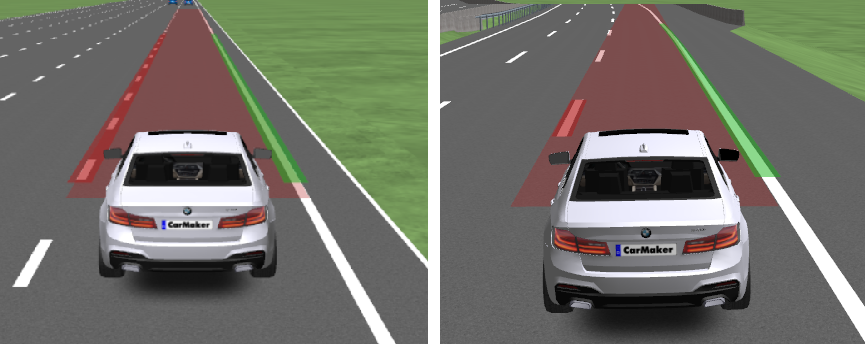}
  \caption{The issue of the lane detection for different road models for the same road network.}
  \label{fig:LKALineDetectionProblem}
\end{figure}

\begin{figure}[h]
  \centering
     \includegraphics[width=0.48\textwidth]{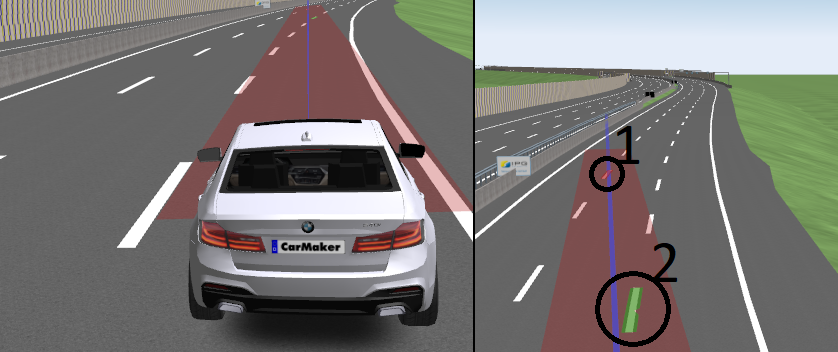}
  \caption{The lane detection issue in the worst case.}
  \label{fig:LKALineDetectionProblemWorstcase}
\end{figure}

In the work of \cite{c2} the lane extension from section \ref{sec:lineDetection} was used for the left and right line marking model for the a LKA, see figure ~\ref{fig:LKALineDetecion}. The implementation works properly for roads which are created directly using the scenario/road editor in IPG CarMaker. However there are certain road network modeling issues which could yield to systematic errors in the lane detection. This is the case for road networks generated by conversion from other road formats like the openDrive format or by converting and modeling networks based on measurements, see \cite{joanneum}. In figure \ref{fig:LKALineDetectionProblem} a false detection behavior is depicted. It can be observed that the ego vehicle, equipped with a lane detection sensor, drives on the road which was measured with different road sensors and extracted in CarMaker native .rd5 file. For this use case, such generated road network yields to issues which are depicted in figure \ref{fig:LKALineDetectionProblem}. In this figure a small but significant difference between two detected line marking can be seen.\\
\hspace{8cm} (a) \hspace{3.7cm} (b)
\\In the right figure ~\ref{fig:LKALineDetectionProblem}(a), the right continuous single line and the left dashed line are detected. Comparing it with the figure \ref{fig:LKALineDetectionProblem}(b) it can be seen that the right single line is detected, but the left dashed line is only partially detected. However, the detection of the left lane marking (LM) differs crucial between figure (a) and (b). In the right one, one segment of the dashed line is covered in red which visualizes that only this piece of the line is used as LM for the whole lane. This segment is only about 6m long~\cite{c5}. This problem occurs because every single dash of the line is created as a own separate line-object. Due to the implementation only one line-object per side gets evaluated. This leads to insufficient number of points for the trajectory calculation. The detection failure can even get worse when driving into a right curve. Drawing a straight line (blue line) in longitudinal direction of the car through its middle, the line sensors will always take the nearest line segments to this virtual line on each side and will use those as left and right guiding line. A demonstration of the detection and the worst case situation is shown in figure \ref{fig:LKALineDetectionProblemWorstcase}.
As left LM, the line segment in red (number 1) gets detected. On the right LM the line segment in green (number 2) is recognized. The blue line is straight in longitudinal direction, starting at the central position of the vehicle. The nearest lines  on each side of the road get selected. This behavior is described in the frequently asked question on the website of IPG~\cite{c4} too. This makes the position and lane detection impossible, because the computation takes these segments as whole guiding lines for the current driving lane. The origin of this problem is, that in road files created with the scenario editor powered by IPG, a dashed line is one single line-object, while in the ALP.Lab for example it is build out of  multiple single line-objects. If the dashed line in figure~\ref{fig:LKALineDetectionProblemWorstcase} was drawn properly, the senors would only  detect four line objects in sum and  assign the the first dashed line to the left and the unbroken line to the right correctly to the current driving lane. The published problem described with this scenario is how to recognize the lane correctly using a cloud of unordered points by ignoring the categorization for the side assignment by IPG. 
\subsection{Advanced Lane Detection Model}
\label{section:lineDetection}
As a solution for the line detection issue from section \ref{section:Problem} the Advanced Lane Detection Model (ALDM) is introduced. The line sensor detects internally multiple lines, even though just one is visualized in CarMaker. The C code interface is used to read the array containing the list of points of measured and valid points for the LM on the street. The structure of the variable for accessing the extended point list provided by the line sensor is:
\begin{lstlisting}
   LineSensor[m].LLines.L[n].ds[p][q]
   LineSensor[m].RLines.L[n].ds[p][q]
\end{lstlisting}
Index \textit{n} defines the line from which the points are accessed. The new lane detection takes in contrast to the existing one not just one line per side but all detected lines.\\
The line sensor has a range of about 200 m in driving direction and creates a measure point every 2 m. It can detect up to 100 lines which leads to an amount of up to 10000 points according equation (\ref{equation:maxNumPoints}).
\begin{table}[h]
	\centering
    \begin{equation}
        n=\frac{ld\_range}{dx}\cdot n_{lines} = \frac{200m}{2m}\cdot100=10000
        \label{equation:maxNumPoints}
    \end{equation}
    \begin{tabular}{lll}
        $ld\_range$&...&range of view for lane detection\\
        $dx$&...&relative distance between two points in x direction\\
        $n_{lines}$&...&total number of theoretically detectable lines \\
    \end{tabular}
\end{table}
To handle this possible amount of data in real time, the ALDM is implemented in C. The algorithm puts all points together in one amount, independently if they where categorised to the left or right side (by figure \ref{fig:LKALineDetectionProblemWorstcase} it is proven that this information may not be valid). Starting with the lowest x-coordinate, which is the nearest point to the car in longitudinal direction, the algorithm computes the next point which has the highest possibility to belong to the same line. Therefore a few points have to be "guessed" as starting condition.
To work properly the data points have to fulfill following conditions:
\begin{itemize}
    \item For each guiding line the first three points muss be guessed right as start points.
    \item The start points for the algorithm must not be too far away from the ego vehicle, this limit is set to 18 m + 5.52m=23.52m. \textbf{Remark}: 5.52m is the offset based on the field of view of the sensor and the nearest detectable point.
    \item The distance between two points of the same line must not be too far apart from each other. This limit is also set to 18 m. \textbf{Remark}: The allowed distance between two point is defined with 18 m because a single segment of a dashed line has to have a length of 6 m and the gap between one segment to the next segment hast to be 12 m according to Rechtsvorschrift fur Bodenmarkierungsverordnung §5 section 2 (see \cite{c5}) at usual road conditions. For this reason there must be at least one line segment within 18 m and so a measured point as data.
    \item The preview distance must be at least 60 m.
\end{itemize}\par
\begin{figure}[h]
	\centering
	\begin{tikzpicture}
	\begin{axis}[
	xlabel={lateral distance in m},
	ylabel={longitudinal distance in m},
	xmin=-6, xmax=6,
	ymin=0, ymax=100,
	yticklabels=\empty,
	legend pos=north west,
	%ymajorgrids=true,
	%grid style=dashed,
	nodes near coords=$\pgfplotspointmeta$,
	nodes near coords align={left},
	point meta=explicit symbolic,
	%nodes near coords style={font=\tiny},
	width=0.48\textwidth
	]

	\addplot[
	only marks,
	color=blue,
	mark=+,
	]table[meta index=2] {figures/dataForVisualizationoftheworkingprincipleoftheALD.txt} ;
	
	\end{axis}
	\end{tikzpicture}
	
	\caption{Visualization of the working principle of the ALDM.}
	\label{fig:startPointPic}
\end{figure}
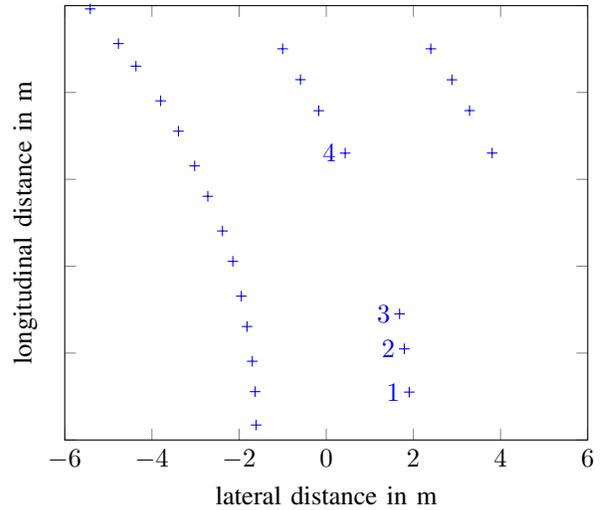
\begin{figure}[ht]
  \centering
     \includegraphics[width=0.48\textwidth]{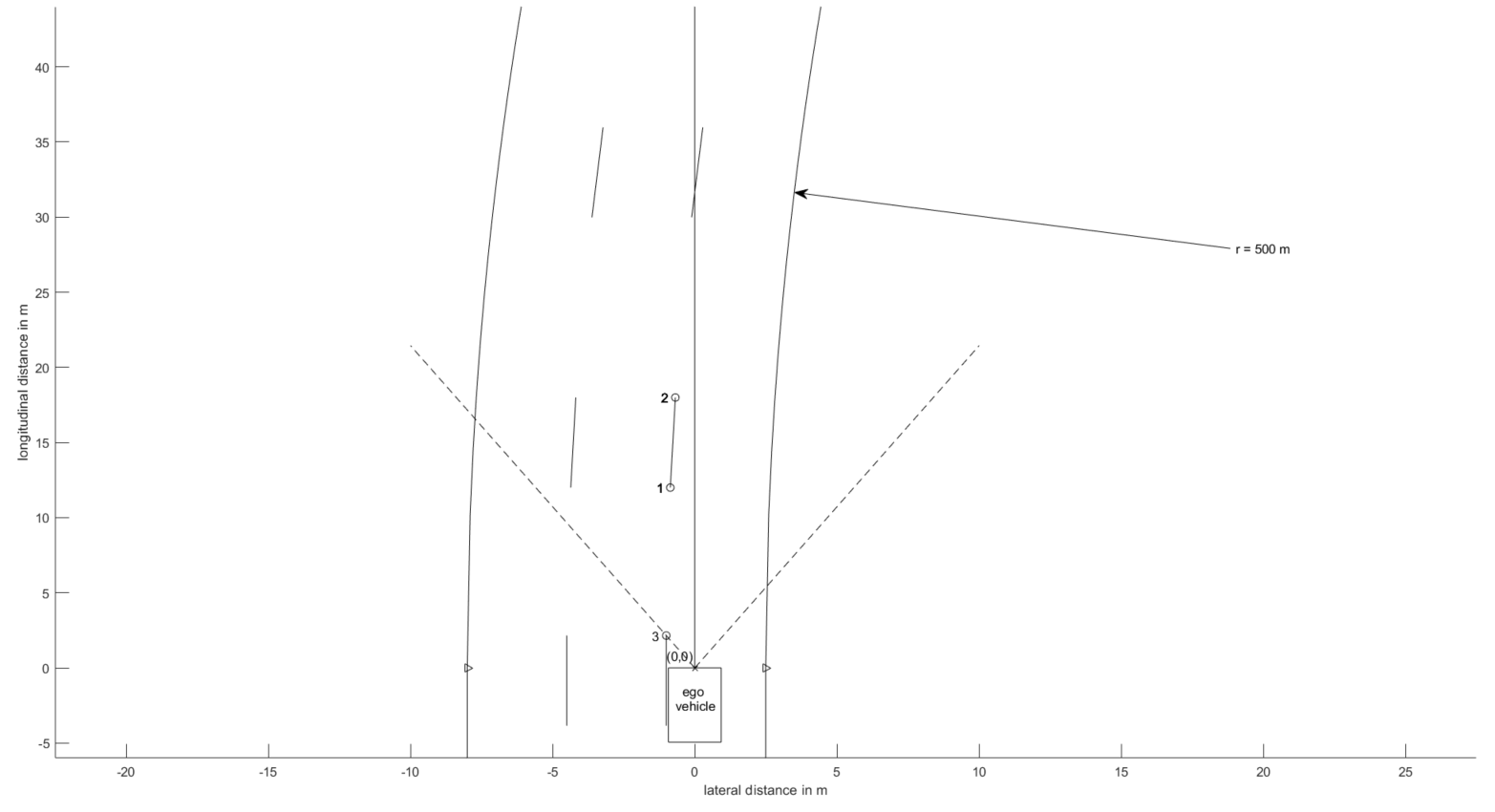}
  \caption{ALDM start conditions in the worst case scenario.}
  \label{fig:advancedLineDetectionStartCondition}
\end{figure}

As start condition the three points with the lowest absolute value of the y (lateral) coordinate on each side are used, these points must be at least 1.8 m apart from each other. Looking at figure \ref{fig:startPointPic}, where the ego vehicle is placed in the origin (0$\vert$0), the start condition for the right line are points "1", "2" and "3". Because of the fact that a dash has a length of about 6 m and the line sensor perceives a point about every 2m, there are at least three point at one line segment, as for all segments in orange and yellow in figure \ref{fig:startPointPic}. So it is sufficient that in the worst case situation at least the nearest line segment of a dashed line is detected on the right side. In this case it is possible to calculate or provide three points for the beginning of the LM determination with ALDM. This worst case is shown by figure \ref{fig:advancedLineDetectionStartCondition}. At the beginning of the strongest allowed curve to right, the car is at the leftest  acceptable position\footnote{The leftmost allowed position is defined as left as possible as long as the tires do not touch the guiding line on the left.} at the current lane.
The "Forschungsgesellschaft Stra{\ss}e- Schiene- Verkehr" defines in "RVS 03.03.23 Linienf\"uhrung und Trassierung" \cite{c6} defines the minimum and recommended curve radius on a highway in Austria as shown in Table \ref{table:RVS030323}. For the LKA the minimum curve radius was assumed with 500 m, this is sufficient for an autopilot driving under normal condition.\par
\begin{table}[ht]
\centering
\begin{tabular}{ |c|c|c|c| } 
    \hline
        \textbf{$R_{empf}$[m]}&\textbf{$V_E$[km/h]}&\textbf{$R_{min}$[m]}\\\hline
        \multirow{4}{4em}{$\geq 1000$} & 130 & 800 \\ 
        & 120 & 600 \\ 
        & 110 & 500 \\ 
        & 100 & 400 \\ 
        \hline
    \end{tabular}
    
    \begin{tabular}{lll}
        $R_{empf}$&...&minimum recommended curve radius\\
        $V_E$&...&velocity at the entry point of the curve\\
        $R_{min}$&...&corner case minimum required curve radius \\
    \end{tabular}
    \caption{Minimum curve radius in Austria}
    \label{table:RVS030323}
\end{table}
The worst longitudinal position of the car is chosen right next to a line segment on the road, so that the line segment is still not in the field of view of the line sensor, out of this the distance to the next line segment is as big as possible. In figure \ref{fig:advancedLineDetectionStartCondition} the ego vehicle is placed right at the beginning of the curve. Its starting point is marked with two triangles on the left and right side of the street. The origin of the figure (0,0) is marked with a cross and it is defined by the position of the line sensor itself. The starting and ending point of the nearest detected line segment to the car is labeled with "1" and "2". Label "3" is the end of the line segment next to the car, which still gets not detected because it is out of the field of view of the sensor.
By proving that both, the start (point "1") and the end (point "2") point, are even in the worst case scenario on the left side perceived, it is induced that the whole segment is on the left side. \par%The prove can be seen in the appendix \ref{Appendix1}.\par
Using this start condition the ALDM continuous with the following steps:
\begin{enumerate}
    \item Calculating a quadratic function (see equation \ref{equation:qudraticfunCalcMat}) as trend of the line, using the previous three points. Looking at figure \ref{fig:startPointPic2} the start points "1","2" and "3" are taken and the yellow continuous line is calculated.
    \item Searching for the next point in a longitudinal range of 18 m by looking for the smallest difference of the y coordinate to the quadratic function. For the next iteration in figure \ref{fig:startPointPic2} the point "4" is found, "2", "3" and "4" is again used for calculation of the purple dashed curve.
    \item Repeating this until the senors range of view is reached. 
\end{enumerate}
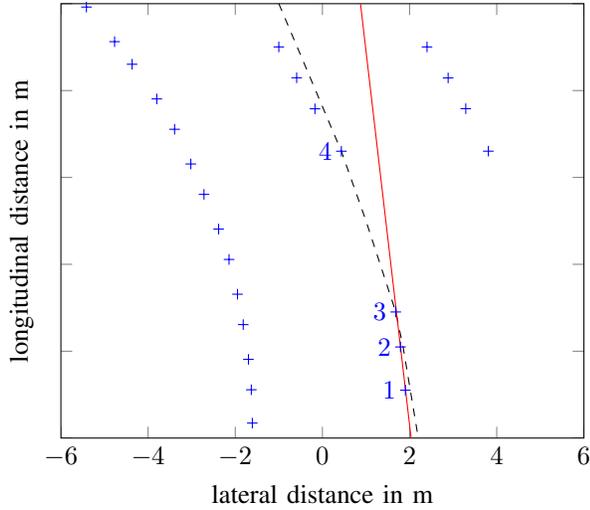
\begin{figure}[ht]
	\centering
	\begin{tikzpicture}
	\begin{axis}[
	xlabel={lateral distance in m},
	ylabel={longitudinal distance in m},
	xmin=-6, xmax=6,
	ymin=0, ymax=100,
	yticklabels=\empty,
	legend pos=north west,
	%ymajorgrids=true,
	%grid style=dashed,
	nodes near coords=$\pgfplotspointmeta$,
	nodes near coords align={left},
	point meta=explicit symbolic,
	%nodes near coords style={font=\tiny},
	width=0.48\textwidth
	]

	\addplot[
	only marks,
	color=blue,
	mark=+,
	]table[meta index=2] {figures/dataForVisualizationoftheworkingprincipleoftheALD.txt} ;
	
	\addplot[
	%only marks,
	color=black,
	%mark=square,
	%mark=+,
	dashed,
	]coordinates {(-1, 100 ) (0.434470, 66.032104 )(1,  50) (1.682679,  29.031204)  (1.851096,  20.971346) (2.2,  0)};
	\addplot[
	%only marks,
	color=red,
	%mark=square,
	%mark=+,
	%dashed,
	]coordinates { (2.035,0) (0.873526,100) } ;

	\end{axis}
	\end{tikzpicture}
	\caption{Visualization of the working principle of the ALDM.}
	\label{fig:startPointPic2}
\end{figure}
Let A, B and C be three points (for example "1", "2" and "3" in figure \ref{fig:startPointPic2}) with the coordinates
\begin{itemize}
    \item[-] $A(x_a|y_a)$
    \item[-] $B(x_b|y_b)$
    \item[-] $B(x_c|y_c)$
\end{itemize}
and a quadratic function defined as
\begin{equation*}
    y=ax^2+bx+c
\end{equation*}
The following system of equation can be created:
\begin{align}
    y_a=ax_a^2+bx_a+c\\
    y_b=ax_b^2+bx_b+c\\
    y_c=ax_c^2+bx_c+c
\end{align}
This can be written as a matrix:
\begin{equation}
    \left[
    \begin{array}{ccc}
        x_a^2&x_a&1\\
        x_b^2&x_b&1\\
        x_c^2&x_c&1
    \end{array}
    \right] \cdot\left[
    \begin{array}{c}
        a\\
        b\\
        c
    \end{array}
    \right]=\left[
    \begin{array}{c}
        y_a\\
        y_b\\
        y_c
    \end{array}
    \right]
\end{equation}
With the solution for the matrix:
\begin{align}
    \left[
    \begin{array}{c}
        a\\
        b\\
        c
    \end{array}
    \right]=
    \left[
    \begin{array}{ccc}
        x_a^2&x_a&1\\
        x_b^2&x_b&1\\
        x_c^2&x_c&1
    \end{array}
    \right]^{-1}
    \cdot \left[
    \begin{array}{c}
        y_a\\
        y_b\\
        y_c
    \end{array}
    \right]
    \label{equation:qudraticfunCalcMat}
\end{align}
The inverse of the 3x3 matrix was implemented in C using the adjunct matrix. The results are the coefficients a,b and c describing a quadratic function. The same working principle used for the left and right LM of the current driving lane, is used for the left LM of the lane left to the ego vehicle and the right LM of the lane on the right side. This way, up to three lanes can be detected, the current driving lane, the lane to the left, and to the right, if existing. The crucial requirements for a reliable lane detection are the three starting points with which the algorithm detects the remaining single points of a line. They are chosen in a similar way as for the other guiding lines, with a condition of a minimum distance of about 2 m in lateral direction  to the guiding lines of the current lane to avoid the detection of the same line twice. The C program uses for the calculation all available  points,  this can be up to 100 points per line. However out of performance reasons and because there is no need for so many points, only 13 equally distributed points out of this amount are finally transmitted to the Simulink interface. Each set of 13 point represents one of the four lines, as depicted in figure \ref{fig:AdvancedLineDetectionExample}. The trajectory for the current lane is computed by calculating two trend lines using a function of 3rd grade using the points detected as the left and right LM. To get the driving trajectory, the difference between those two lines is calculated, results see in section \ref{section:LKAResult}.
\section{Simulation Result}
\label{section:LKAResult}
\begin{figure}[h]
	\centering
	\begin{tikzpicture}
	\begin{axis}[
	xlabel={lateral distance in m},
	ylabel={longitudinal distance in m},
	xmin=-15, xmax=11,
	ymin=0, ymax=200,
	legend pos=north west,
	%ymajorgrids=true,
	%grid style=dashed,
	nodes near coords=$\pgfplotspointmeta$,
	nodes near coords align={right},
	point meta=explicit symbolic,
	nodes near coords style={font=\tiny},
	width=0.48\textwidth
	]

	\addplot[
	only marks,
	color=blue,
	mark=+,
	]table [meta index=2] {figures/dataForALDmeasurementraw-data.txt} ;
	
	\end{axis}
	\end{tikzpicture}
	\caption{Visualization of the LM using the ALDM.}
	\label{fig:AdvancedLineDetectionResult}
\end{figure}
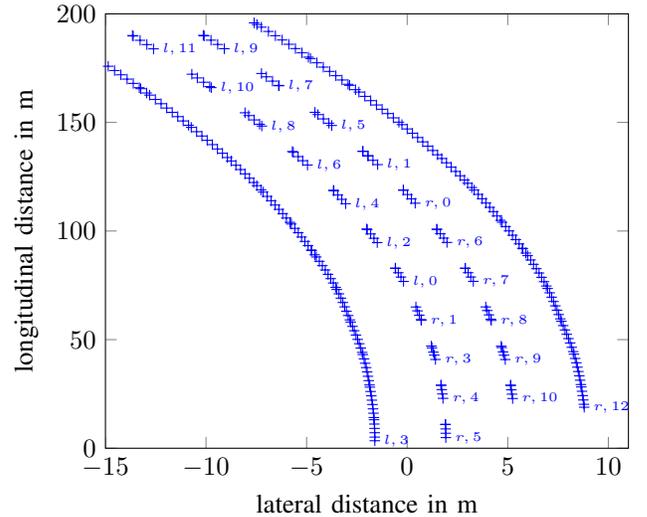

In figure~\ref{fig:AdvancedLineDetectionResult} are all detected LM using the ALDM. The labels are describing if the points were assigned to the left or right side and which line index is used\footnote{Compare index \textit{n} in table~\ref{listing:listingCVariables}}. Marker 2 of picture~\ref{fig:LKALineDetectionProblemWorstcase} is labeled as "right,0"  while  marker 1 is  "left,0".
An  amount of points  which cover the lines painted on the whole street in front of the ego vehicle is shown by the figure \ref{fig:AdvancedLineDetectionResult}. These values are calculated by the ALDM as described in \ref{sec:lineDetection}. A reduced amount of points of the left and right guiding line of the current driving lane  gets transmitted to the Simulink simulation.
\begin{figure}[ht]
	\centering
	\begin{tikzpicture}
	\begin{axis}[
	xlabel={lateral distance in m},
	ylabel={longitudinal distance in m},
	xmin=-15, xmax=11,
	ymin=0, ymax=200,
	legend pos=north west,
	%ymajorgrids=true,
	%grid style=dashed,
	nodes near coords=$\pgfplotspointmeta$,
	nodes near coords align={right},
	point meta=explicit symbolic,
	nodes near coords style={font=\tiny},
	width=0.48\textwidth,
	]

	\addplot[
	only marks,
	color=blue,
	mark=o,
	]table{figures/dataL3ForAdvancedLaneDetectionresult.txt} ;
	\addplot[
	only marks,
	color=blue,
	mark=o,
	]table{figures/dataL2ForAdvancedLaneDetectionresult.txt} ;
	\addplot[
	only marks,
	color=blue,
	mark=+,
	]table{figures/dataL1ForAdvancedLaneDetectionresult.txt} ;
	\addplot[
	only marks,
	color=blue,
	mark=+,
	]table{figures/dataL4ForAdvancedLaneDetectionresult.txt} ;
	\addplot[
	%only marks,
	color=red,
	]table{figures/dataL3ForAdvancedLaneDetectionresult.txt} ;
	\addplot[
	%only marks,
	color=red,
	]table{figures/dataL2ForAdvancedLaneDetectionresult.txt} ;
	\addplot[
	%only marks,
	color=green,
	%mark=square,
	%mark=+,
	]table{figures/dataL1ForAdvancedLaneDetectionresult.txt} ;
	\addplot[
	%only marks,
	color=green,
	%mark=square,
	%mark=+,
	]table{figures/dataL4ForAdvancedLaneDetectionresult.txt} ;

	\end{axis}
	\end{tikzpicture}
	\caption{Lane detection points calculated by the ALDM.}
	\label{fig:AdvancedLineDetectionShowingPresentedSituationResult}
\end{figure}
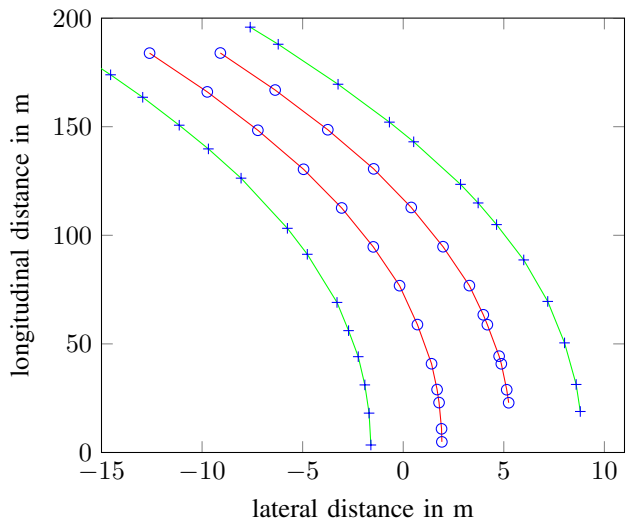
Figure~\ref{fig:AdvancedLineDetectionShowingPresentedSituationResult} reveals the final result of the algorithm. The points marked with blue crosses represent the left line of the lane and the orange circles are the right line, descending from the points in plot~\ref{fig:AdvancedLineDetectionResult}. Both quantities are interpolated, shown in the dashed thin curves in orange and purple. The green thicker dotted line in the middle is the calculated desired driving trajectory.
%%%%%%%%%%%%%%%%%%%%%%%%%%%%%%%%%%%%%%%%%%%%%%%%%%%%%%%%%%%%%%%%%%%%%%%%%%%%%%%%%%%%%%%%%%%%%%%%%%%%%%%%%%%%%%%%%%%%%%%%%%
\label{section:LCAResult}
\begin{figure}[ht]
  \centering
     \includegraphics[width=0.48\textwidth]{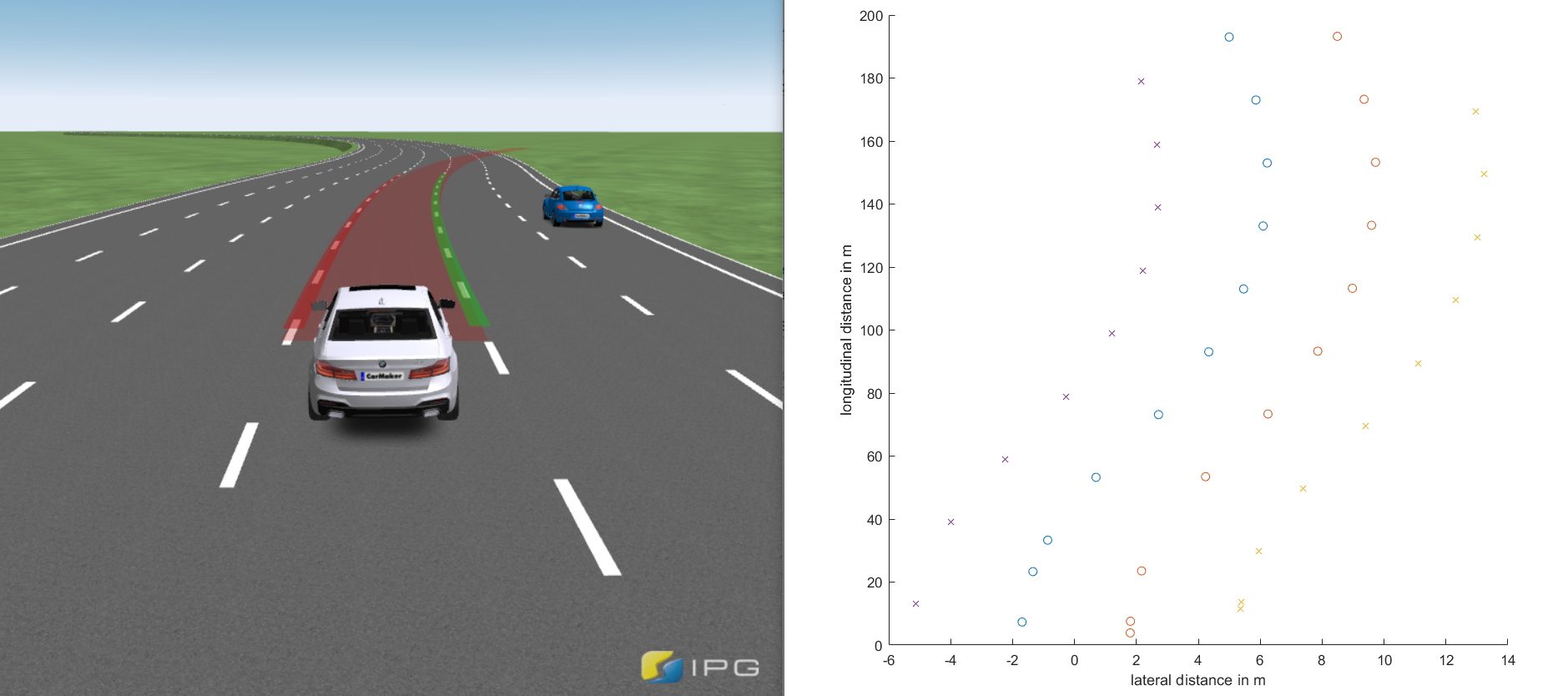}
  \caption{Lane detection points of the ALDM for the LCA.}
  \label{fig:AdvancedLineDetectionExample}
\end{figure}
Figure~\ref{fig:AdvancedLineDetectionExample} shows the ego vehicle  driving on the third lane counted from the right edge of the road as well as the detected lines for this situation on the right side. The origin (0$vert$0) of the right graph is assigned to the position of the car in the middle of the third lane. The points marked with a circle (blue and orange) are the guiding lines of the current lane and the line consisting of crosses (purple and yellow) represent the next line on the left and right side. As seen, the LCA detects the adjacent lanes as well as the current driving lane. The information about the additional lines are  wrapped into a set of points using a vector and are transmitted to the Simulink simulation. The LCA is able to use these information for calculating the trajectory when changing lane as well.

\addtolength{\textheight}{-12cm}   % This command serves to balance the column lengths

%%%%%%%%%%%%%%%%%%%%%%%%%%%%%%%%%%%%%%%%%%%%%%%%%%%%%%%%%%%%%%%%%%%%%%%%%%%%%%%%

\end{document}